\documentclass[10pt,twocolumn,letterpaper]{article}

\usepackage{comment}
\usepackage[normalem]{ulem}
\usepackage{times}
\usepackage{epsfig}
\usepackage{graphicx}
\usepackage{amsmath}
\usepackage{amssymb}
\usepackage[utf8]{inputenc} 
\usepackage[T1]{fontenc}    
\usepackage{url}            
\usepackage{booktabs}       
\usepackage{amsfonts}       
\usepackage{nicefrac}  
\usepackage{microtype} 
\usepackage{bm}
\usepackage{bbm}
\usepackage{times}
\usepackage{authblk}
\usepackage{epsfig}
\usepackage{algorithm}
\usepackage{algpseudocode}
\usepackage{xcolor,colortbl}
\usepackage{graphicx}
\usepackage{color}
\usepackage{stmaryrd}
\usepackage{graphics}
\usepackage{mathptmx}
\usepackage{xspace}
\usepackage{dsfont}
\usepackage{mathtools}
\usepackage{tabularx, multirow}
\usepackage{gensymb}
\usepackage{indentfirst}
\usepackage[pagebackref=true,breaklinks=true,letterpaper=true,colorlinks,bookmarks=false]{hyperref}
\usepackage{cleveref}

\begin{document}

\title{THIN: THrowable Information Networks and Application for Facial Expression Recognition In The Wild}

\author[1]{Est\`ephe Arnaud}
\author[2]{Arnaud Dapogny}
\author[1, 2]{K\'evin Bailly}

\affil[1]{\small Sorbonne Universit\'e, CNRS, Institut des Systèmes Intelligents et de Robotique, ISIR, F-75005 Paris, France}
\affil[2]{\small Datakalab, Paris, France}

\date{}
\maketitle

\begin{abstract}
    For a number of machine learning problems, an exogenous variable can be identified such that it heavily influences the appearance of the different classes, and an ideal classifier should be invariant to this variable. An example of such exogenous variable is identity if facial expression recognition (FER) is considered.
    In this paper, we propose a dual exogenous/endogenous representation. The former captures the exogenous variable whereas the second one models the task at hand (e.g. facial expression). We design a prediction layer that uses a tree-gated deep ensemble conditioned by the exogenous representation.
    We also propose an exogenous dispelling loss to remove the exogenous information from the endogenous representation. Thus, the exogenous information is used two times in a throwable fashion, first as a conditioning variable for the target task, and second to create invariance within the endogenous representation. We call this method THIN, standing for THrowable Information Networks. We experimentally validate THIN in several contexts where an exogenous information can be identified, such as digit recognition under large rotations and shape recognition at multiple scales. We also apply it to FER with identity as the exogenous variable. We demonstrate that THIN significantly outperforms state-of-the-art approaches on several challenging datasets. 
\end{abstract}
    
\section{Introduction}\label{sec:introduction}
Facial expression recognition (FER) is an active computer vision field. It consists in estimating the perceived emotional state of a person based on facial cues, with applications in human-computer interaction \cite{cowie2001emotion}\cite{bartlett2003real}, virtual/augmented reality \cite{bekele2013understanding}\cite{chen2015augmented}, advanced driver assistance systems \cite{assari2011driver}\cite{jabon2010facial}, education \cite{kapoor2007automatic}, entertainment \cite{lankes2008facial} and healthcare \cite{dapogny2018jemime}.

Currently, deep learning techniques have led to a significant advance in FER, allowing to jointly learn a representation, which we call an \textbf{endogenous} representation, and a predictor based on this representation. However, the appearance of the face can be heavily influenced by an \textbf{exogenous} variable (e.g. identity-related information, head pose, occlusions) from which the task prediction shall ideally be invariant.

\begin{figure}
    \centering
    \includegraphics[width=\linewidth]{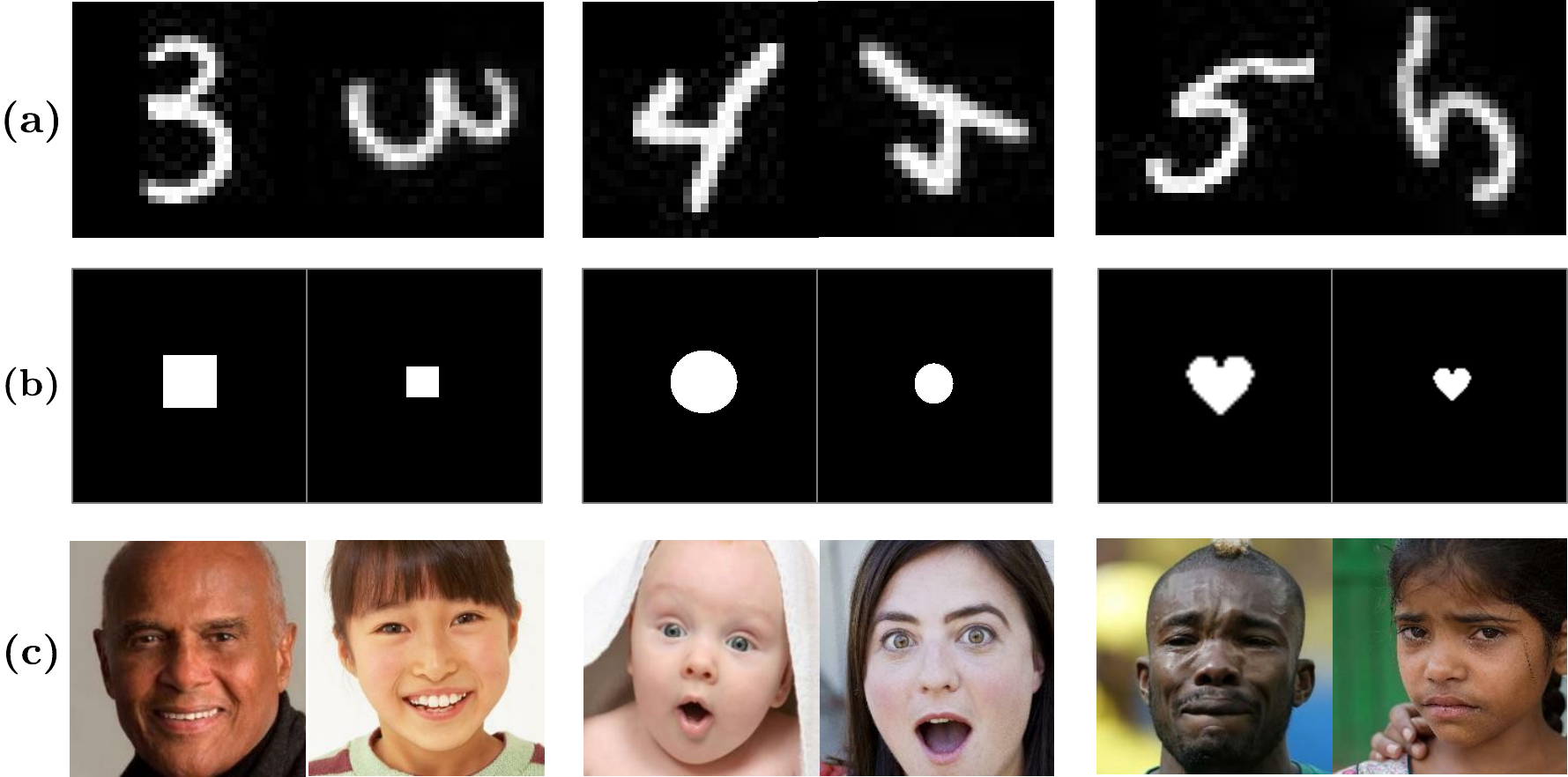}
    \caption{Examples of tasks where an exogenous variable can be identified. \textbf{(a):} digit recognition with rotation as the exogenous variable. \textbf{(b):} shape recognition with scale. \textbf{(c):} FER with identity as the exogenous variable.}
    \label{dsv}
\end{figure}

Figure \ref{dsv} shows three different tasks where such an exogenous variable can be identified. In Figure \ref{dsv} - (a), the digit appearance is dramatically influenced by its rotation. In the same vein, in Figure \ref{dsv} - (b), the recognition of the shape of the object is strongly influenced by its scale. Finally, for more complex tasks like FER as illustrated on Figure \ref{dsv} - (c), a subject morphology dramatically affects its expressive face appearance. Thus, in such a case, identifying the exogenous variable and using it to condition the model can greatly help the task prediction.

\begin{figure*}[!t]
  \centering
  \includegraphics[width=0.95\linewidth]{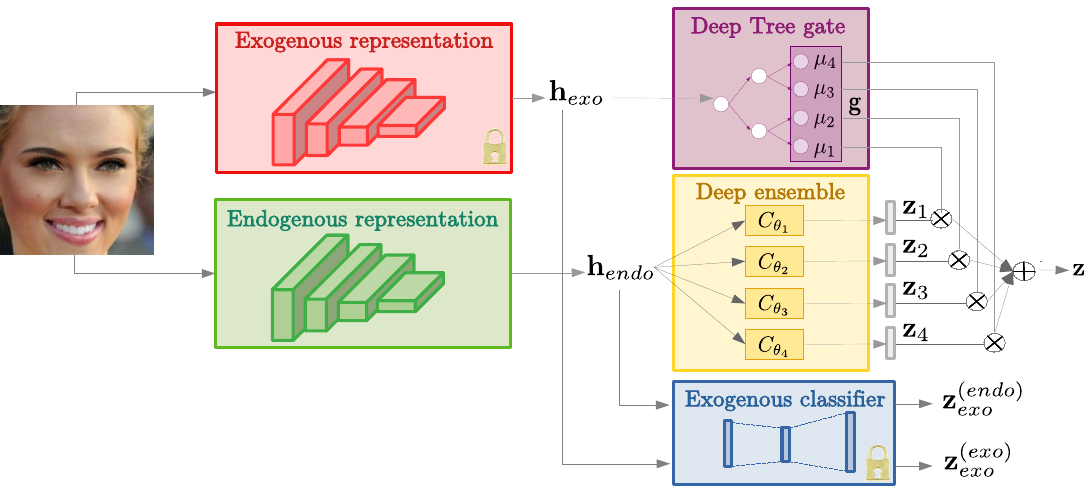}
  \caption{Overview of THIN with 4 classifiers in the deep ensemble. The exogenous variable is used two times, in a throwable fashion, to enhance the prediction. For instance, for FER with identity as the exogenous variable, FER is conditioned by the identity representation $h_{exo}$ as estimated by a first network. Furthermore, exogenous information is removed from the endogenous representation $h_{endo}$ by applying our exogenous dispelling loss. The padlock symbol illustrates the fact that both the exogenous representation and classifier are pretrained beforehand, and frozen during the joint training of the endogenous representation, deep tree gate and ensemble.}
  \label{bigpicture2}
\end{figure*}

However, at the same time, from the perspective of a prediction system, our target task (e.g. facial expressions) shall be predicted regardless of the variations of the exogenous variable (e.g. subject identity). We therefore argue that a representation that is relevant to our task (which we call an endogenous representation) shall contain as little information about the exogenous variable as possible.

To sum it up, in such a case, this exogenous variable is an important source of variation in the data and, at the same time, an information from which the output of a predictor shall be rendered as invariant as possible. For this reason, we propose to use separate exogenous and endogenous representations. The exogenous representation is used two times, in a throwable fashion: (a) as a conditioning variable for the target task prediction, and (b) in the frame of a dispelling loss to explicitly remove the exogenous information from the endogenous representation.

In order to enforce (a) we propose to condition the target task prediction according to a representation related to the exogenous variable. Adapted to FER for instance, identity-related information (which encompasses particular morphological traits, gender, age or ethnicity) will be used to specialize the prediction accordingly. Note that, to a certain extent, human understanding of facial expressions works similarly: if we know that a subject has frowned eyebrows or dimples will greatly help to make a decision whether that person expresses anger or a subtle smile. In order to implement that, we propose to use a committee of weak predictors instead of a single strong predictor to generate a set of diverse possible predictions. Also, we adaptively select the most relevant weak predictors depending on the exogenous representation. This adaptive weak predictor mixture can be learned jointly by using differentiable tree gates. Furthermore, it allows to learn a partition in the exogenous representation space, upon which the weak predictors can specialize on.
This deep ensemble method then allows to obtain well-suited specialist weak predictors for each cluster in the exogenous representation space rather than having a single predictor for all cases, thus increasing the overall robustness to variations of the exogenous variable.
In order to enforce (b), we propose a novel dispelling loss to explicitly remove the exogenous information from the endogenous representation, by leveraging an exogenous classifier (blue box on Figure \ref{bigpicture2}). The upstream specialist weak predictors (yellow box on Figure \ref{bigpicture2}), the gates that learns the adaptive mixture (purple box) based on the exogenous representation (red box) and the endogenous representation layer (green box) are all trained jointly in an end-to-end manner.

This method is quite general and can be applied to a variety of problems, where an exogenous variable can be identified. We call this method THIN, standing for THrowable Information networks. In particular, we adapt THIN to several contexts and tasks where an exogenous variable can be identified: digit recognition on MNIST dataset augmented with random rotation (with rotation as the exogenous variable), and shape recognition on dSprites dataset (with the shape scale as the exogenous variable). We also apply THIN to real-world FER databases with identity as the exogenous variable. To sum it up, the contributions of this paper are thus three-folds:

\begin{itemize}
    \item From an architectural standpoint, we propose a new adaptive deep architecture using a gating variable exogenous to the target task to better condition it and learn weak classifiers more robust to these exogenous variations.
    \item From a learning standpoint, we propose a new training loss encouraging to remove the exogenous information from the endogenous representation, further improving the overall learning algorithm and the robustness of the weak classifiers to exogenous variations.
    \item From an experimental standpoint, we apply our approach to multiple predictive tasks (e.g. digit recognition, shape recognition as well as FER). In particular, we show that our THIN model significantly outperforms the state-of-the-art methods on today's most challenging FER databases.
\end{itemize}

The paper is organized as follows. In Section \ref{sec:related_work}, we review related work on deep ensemble methods and learning disentangled representations, as well as deep methods for FER, and in particular Identity-based FER methods. In Section \ref{sec:framework}, we provide an overview of our THIN model. We then demonstrate empirically the interest of our approach for multiple classification tasks: digit and shape recognition in Section \ref{sec:experiments_mnist_dsprites}, and FER in \textcolor{black}{\cref{emo_datasets,implemdetails,ablation,compsota,modelintro}}. Finally, in Section \ref{sec:conclusion}, we give some concluding remarks about our approach and discuss future work.

\section{Related work}\label{sec:related_work}
Relatively to the above mentioned contributions, we review existing work on deep learning for FER, with a particular emphasis on methods that leverage identity information. We also review existing methods for deep ensemble learning, and more specifically conditioning in deep ensembles. Finally, we review recent related work on disentangling representations in deep learning.

\subsection{Deep learning for FER} 
A simple historic application of deep learning techniques to FER is to use classical pretrained networks (e.g. AlexNet, VGG, ResNet) and fine-tune them for FER \cite{mollahosseini2016going}\cite{mollahosseini2016facial}. Since then, multiple improvements have been introduced. This encompasses for instance architectural changes \cite{hasani2019bounded}\cite{Acharya_2018_CVPR_Workshops}, the design of a specific loss function \cite{li2019separate} or the use of attention mechanisms to adaptively capture the importance of specific facial regions \cite{wang2019region}\cite{Li2018PatchGatedCF}.

Recently, some other methods aim to enforce robustness by explicitly integrating identity-related information within the FER pipeline. This exogenous information  is usually handled either using metric learning techniques \cite{meng2017identity}\cite{liu2017adaptive}\cite{liu2019hard} or adversarial training \cite{zhang2018identity}\cite{wang2019identity}. 
As such, Meng et al. \cite{meng2017identity} propose to design a contrastive loss, which consists in jointly learning similarity metrics related to facial expression and identity from pairs of samples annotated with either information.
Liu et al. \cite{liu2017adaptive} proposed to use a triplet loss, by jointly decreasing the distance between an anchor and a positive sample (from the same FE class and different identities) and increasing the distance between the anchor and a negative sample (from a different FE class or same identity). In a later work \cite{liu2019hard} they extended this approach by using generative adversarial networks to provide synthetic neutral faces for each example. However, the training is difficult to tune, computationally expensive, and the generator may fail to produce photo-realistic face images where the identity is preserved. Zhang et al. \cite{zhang2018identity} trained a shared representation layer by minimizing an facial expression recognition loss while maximizing an identity recognition loss with identity-related adversarial training. Whang et al. \cite{wang2019identity} use adversarial learning for identity and pose-robust facial expression recognition. However, these methods can only be trained on datasets annotated in terms of both FEs and identity. This is not the case of in-the-wild datasets \cite{li2017reliable}\cite{mollahosseini2017affectnet}, which contain web-scrapped images without identity annotation. 

By contrast, the proposed method doesn't require any such information, as a separate deep network can be trained on a separate face recognition database beforehand. This exogenous (identity) network will be used to condition a deep ensemble to produce specialists networks, which is another way to deal with the large intra-class variability in FER datasets. 

\subsection{Deep ensemble methods}

Ensemble learning is known to be an effective way to increase the prediction robustness and accuracy. Hence, different strategies have been successfully proposed in the literature to train and combine deep neural networks. The most straightforward way consists in averaging the output of the DNN ensemble. To be effective, the diversity of the ensemble has to be ensured, for example, by initializing each DNN differently \cite{wen2017ensemble}, by updating parameters of only one model at each step of the training \cite{kontschieder2015deep}, by diversifying the model architecture \cite{yvinec2020deesco}, or by saving the model parameters at several local minima during training \cite{HuangLP0HW17}.

Rather than simply averaging the predictions of the classifier ensemble, other approaches have integrated a gating network within the deep ensemble to adaptively weight the predictions. The combination depends on the input upon which downstream deep ensemble is inferred. The idea was firstly introduced by Jacobs et al. \cite{jacobs1991adaptive} and more recently taken up by Eigen \textit{et al.} \cite{eigen2013learning}. The latter
designed a Mixture-of-Experts layer, where both a set of expert sub-networks and softmax gate (mixture) are learned jointly. This allows to learn both the specialists, and the corresponding mixture weights in an end-to-end manner, resulting in more optimal representations and predictions.
In the same vein, Shazeer \textit{et al.} \cite{shazeer2017outrageously} introduce sparsity in MoE layer, which improves the inference times while maintaining the expressiveness of the network. Arnaud et al. \cite{ArnaudFG2019} use a tree-structured gate that acts as a hierarchical soft partitioning of the input space and they successfully apply this method for face alignment. In \cite{ArnaudTBIOM2019}, they show that the partitioning of the input space can be conditioned by a high level semantic variable such as head pose. This demonstrates that the use of an explicit exogenous variable that is responsible for strong variations in the data distribution, can improve the robustness and accuracy of the result. 

\subsection{Learning disentangled representations}
There are a multitude of methods in the literature for learning disentangled representations. Traditionally, these approaches use an autoencoders, assuming that the representation extracted from the encoder factorizes all the sources of variation. These disentangling methods then aim at assigning a physical interpretation to each dimension of this latent representation. These methods differ in their training procedure: supervised \cite{hinton2011transforming}\cite{cheung2014discovering}\cite{kulkarni2015deep}\cite{louizos2016variational}\cite{mathieu2016disentangling} or unsupervised learning \cite{denton2017unsupervised}\cite{hsu2017unsupervised}\cite{kim2018disentangling}. 

A first approach \cite{kulkarni2015deep} consists in explicitly varying a single source of variation for each training batch so as to update the weights only for the associated embedding. But this technique requires to have access to these sources of variations and to be able to generate them synthetically.
Other approaches \cite{hinton2011transforming}\cite{hsu2017unsupervised} still require the variable values for each sources of variations for each training sample (\textit{e.g.} translation, rotation, \textcolor{black}{etc}) to separate the different associated embeddings. Thus, this implies that the train set shall be annotated both in terms of the exogenous and endogenous variables. Conversely, in this work, we can train on separate datasets as we use a pretrained network to model the exogenous representation.
Finally, a last approach consists in designing specific losses in order to make the embeddings complementary to each other, either by applying a penalty term for orthogonalizing the embeddings \cite{cheung2014discovering}, or by means of an adversarial loss where a decoder is given the embeddings generated by two different labeled samples \cite{mathieu2016disentangling}. In such a case, the training is difficult to tune and computationally expensive, whereas our approach uses only the exogenous variable and aims to semantically orthogonalize the dual representations. We adopt an approach similar to \cite{cheung2014discovering} for its simplicity and efficiency. However, rather than encouraging the orthogonality among latent dimensions of the raw representations, we aim to orthogonalize them in a space of a higher semantic level, \textit{i.e.} the exogenous representation.

\section{Framework overview}\label{sec:framework}

\begin{figure*}[!t]
  \centering
  \includegraphics[width=\linewidth]{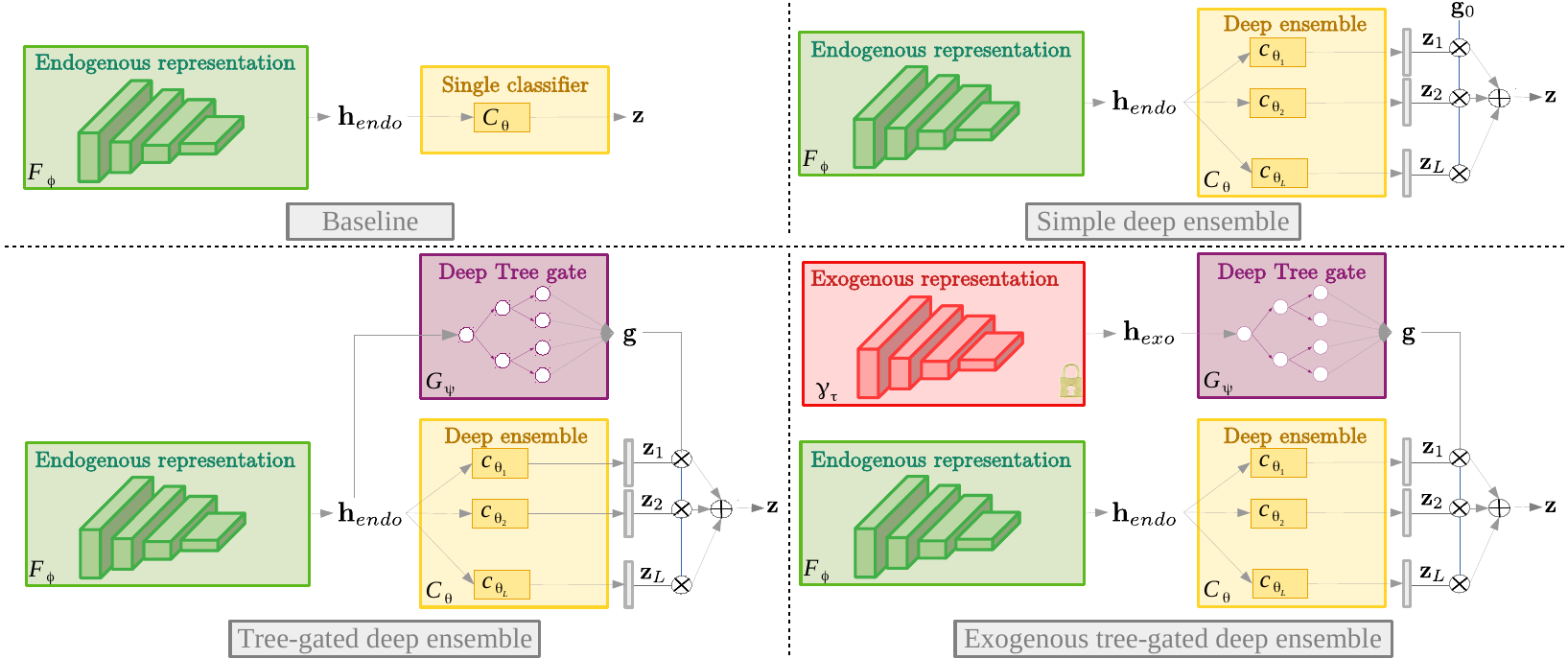}
  \caption{Architectures overview. \textbf{Baseline} (top-left): a simple baseline, composed of a stack of representation and classification layers. \textbf{Single deep ensemble} (top-right): the classification layer is represented an average of several smaller classifiers. \textbf{Tree-gated deep ensemble} (bottom-left): instead of merely averaging the classifiers we learn the mixture weights using a deep differentiable tree. \textbf{Exogenous tree-gated deep ensemble} (bottom-right): we use a second representation network related to the exogenous representation, which appears as a better conditioning variable.}
  \label{archicomparison}
\end{figure*}

In this section, we introduce our THIN model. First, in Section \ref{sec:exo}, we discuss how we model the exogenous information with a deep network. Second, in Section \ref{sec:deepensemble}, we describe how to use this exogenous representation within a deep ensemble architecture to condition the prediction. Starting from a simple deep ensemble as baseline, we discuss its shortcomings and show how we can design a more efficient method by adaptively weighting the predictions of the deep ensemble and by leveraging the exogenous representation. Third, in Section \ref{sec:disentangling} we introduce our exogenous dispelling loss to remove the exogenous information from the endogenous representation. Finally, in Section \ref{sec:difer} we illustrate our method by applying it to FER.

\subsection{Modeling the exogenous information}\label{sec:exo}

To model the exogenous variable, we first train a deep neural network $c_{\sigma}(\gamma_{\tau}(I))$ where $c_{\sigma}$ is the exogenous predictor with parameters $\sigma$ and using the exogenous representation extracted by the network $\gamma_{\tau}: I \in \mathbb{D} \mapsto \textbf{h}_{exo} \in \mathbb{H}_{exo}$ with parameters $\tau$ and based on the image $I$.
Once trained, the parameters $\tau$ are kept frozen so that the exogenous representation $\textbf{h}_{exo}$ can then be integrated within another deep neural network that predicts the task conditionally to it.
Under the condition that the exogenous variable can be (1) identified as an important source of variation in the data, and (2) modeled with sufficient accuracy, we argue that its representation is an ideal variable to condition task prediction, as we will discuss in the upcoming section.

\subsection{Using exogenous information to condition task prediction}\label{sec:deepensemble}

In this Section, we describe how we explicitly incorporate the dependency between the exogenous representation and the task prediction. In particular, we start by defining a simple baseline model then incrementally introduce more efficient methods. Figure \ref{archicomparison} provides an illustration of these architectures.

\paragraph*{A baseline deep approach} as illustrated on Figure \ref{archicomparison} (top-left corner), a baseline deep architecture contains a stack of representation and classification layers. The representation layer $F_\phi: I \in \mathbb{D} \mapsto \textbf{h}_{endo} \in \mathbb{H}_{endo}$ with parameters $\phi$ that takes as input an image $I$ and extracts an embedding vector $\textbf{h}_{endo}$ in the representation space $\mathbb{H}_{endo}$ endogenous to the task. This representation then goes through a classification head denoted $\mathcal{C}_\Theta: \textbf{h}_{endo} \in \mathbb{H}_{endo} \mapsto \textbf{z} \in \mathbb{R}^{K}$ with $\textbf{z}$ an output logit vector and $K$ the number of classes. 
The final prediction outputted by such model is $ \hat{\textbf{y}} \in [0, 1]^K$ defined by applying a softmax activation to $\textbf{z}$. Such baseline network is traditionally trained by gradient backpropagation of a cross-entropy loss between $\hat{\textbf{y}}$ and $\textbf{y}^*$, the corresponding one-hot encoded ground truth vector.

\paragraph*{A simple deep ensemble approach} As ensemble methods are known to improve the generalization capacity \cite{dietterich2000ensemble} and therefore the robustness to large variations, we propose to use a committee of $L$ smaller deep fully-connected networks $c_{\Theta_l}$ instead of using a single large classifier, as it is illustrated on Figure \ref{archicomparison} (top-right corner): 
\begin{equation}
  \mathcal{C}_{\Theta} = \{c_{\Theta_l}\}_{l \in {1 \dots L}}    
\end{equation}
where each expert $c_{\Theta_l}$ outputs a vector logits $\textbf{z}^l \in \mathbb{R}^K$, and $\Theta = \cup_{l=1}^L \Theta_l$ the set of learnable parameters for the whole ensemble. We note $\textbf{Z} \in \mathbb{R}^{K \times L}$ the column matrix containing all the logits vectors outputted by $\mathcal{C}_{\Theta}$:

\begin{equation}
  \textbf{Z} = [ \textbf{z}_1, \dots, \textbf{z}_{L}]
\end{equation}

In case of such a simple deep ensemble, the logits of the classification layer can then be written as the average of each weak predictor, \textit{i.e.}  $\textbf{z} = \textbf{Z} .\textbf{g}$ with $\textbf{g}=\textbf{g}_0=[1/L \dots 1/L]^t$.

\paragraph*{Tree-gated deep ensemble} (Figure \ref{archicomparison}, bottom-left corner) in order to provide a better mixture of the weak predictors we learn a mapping function between $I$ (or, generally speaking, any representation $\textbf{h}_g=\gamma(I)$) and the weights of the mixture $G_\psi: \textbf{h}_g \in \mathbb{H}_g \mapsto \textbf{g} \in [0,1]^L$ with parameters $\psi$ and s.t. $\sum_{l=1}^L \textbf{g}_l = 1$.
The representation $\textbf{h}_g$ of a single image $I$ in the gating space $\mathbb{H}_g$ then allows to condition the use of the committee by emphasizing certain classifiers rather than others depending on the region where $\textbf{h}_g$ is located.

To model $G_\psi$, we propose to use a tree-gate, as in \cite{ArnaudFG2019}. A tree gate allows to learn a hierarchical partition $\cup_{l=1}^L \mathbb{H}_{g,l}$ of the gating space $\mathbb{H}_g$, where coincidentally each classifier $c_{\Theta_l}$ is specialized on a specific region $\mathbb{H}_{g,l}$ (where each $\mathbb{H}_{g, l}$ is a hyper-rectangle).
If $\textbf{h}_{g}$ is in region $\mathbb{H}_{g, l}$, the specialized weak classifier $c_{\Theta_l}$ has the most weight, and others from the most remote regions have the least weights. 
Furthermore, by the use of tree-structured gates, we can then hierarchically cluster the specialized weak classifiers over large regions of the gating space $\mathbb{H}_{g}$.
In early work \cite{ArnaudFG2019}\cite{ArnaudTBIOM2019}, we showed that learning $\mathbb{H}_{g}$ as a hierarchical partition allows to be significantly more robust while using few expert classifiers for the final prediction.
Using a tree gate thus allows to learn a better specialization of each weak classifier, and a better recombination thereof.

To enable end-to-end joint learning of the predictor and gate, we use a neural tree \cite{kontschieder2015deep} that is composed of subsequent soft, probabilistic routing functions $d_n$, that represents the probability to reach the left child of node $n$. Formally, $d_n$ is defined as a single sigmoid unit:
\begin{equation}
  d_n(\textbf{h}_{g}) = \sigma(\textbf{w}_n.\textbf{h}_{g} + b_n)
\end{equation}

with $\Theta_g = \{\textbf{w}_n, b_n\}_{n \in \mathcal{N}}$ the learnable parameters. For an input $\textbf{h}_{g}$, the probability $\mu_l$ to reach a leaf $l$ is computed as a product of the successive activations $d_n$ down the whole tree:
\begin{equation}
  \mu_l(\textbf{h}_{g}) = \prod_{n \in \mathcal{N}} d_n(\textbf{h}_{g})^{\mathds{1}_{l \swarrow n}}(1 - d_n(\textbf{h}_{g}))^{\mathds{1}_{l \searrow n}}  
\end{equation}
where $l \swarrow n$ is true if $l$ belongs to the left subtree of node $n$, and $l \searrow n$ is true if $l$ belongs to the right subtree. We define our tree-gates as the concatenation of the $L=2^\mathcal{D}$ leaves probabilities of a single neural tree of depth $\mathcal{D}$:
\begin{equation}
  \textbf{g} =  [\mu_{1}(\textbf{h}_{g}) \dots \mu_{L}(\textbf{h}_{g})]
\end{equation}

The hyperplane parameters $\Theta_g$ thus delimit the regions of $\mathbb{H}_g$ on which each classifier of the committee is specialized. 

With this in mind, we have yet to chose the nature of the gating space $\mathbb{H}_g$. A baseline solution consists of simply using the endogenous representation $\mathbb{H}_{g} = \mathbb{H}_{endo}$. As we will show, this generally leads to an increase in accuracy as compared to the baselines architectures. 
However, as opposed to such naive approach, we argue that using different representations for expert and gating respectively allows to better specialize the weak classifiers.

\paragraph*{Exogenous tree-gated deep ensemble} the predictive capacity of the tree-gated deep ensemble model is thus linked to \textbf{(a)} the ability of the gate to identify the correct space regions given an image $I$ and \textbf{(b)} the accuracy of the corresponding weak classifiers.
If $\mathbb{H}_g=\mathbb{H}_{endo}$, \textbf{(a)} and \textbf{(b)} are very much related, because classification is inferred on the endogenous representation thus already contains discriminative information. Thus, if relevant clusters can be formed in $\mathbb{H}_{endo}$, a simple predictor on top of the corresponding representation shall have high-end accuracy. Conversely, if an example is hard to predict, it is also likely to be fall into an incorrect cluster in $\mathbb{H}_{endo}$. Hence, it is also likely to be misclassified by the tree-gated deep ensemble.

\textit{A contrario}, a much more interesting case would be to set $\mathbb{H}_{g}=\mathbb{H}_{exo}$ (Figure \ref{archicomparison} - bottom-right corner). We call this the exogenous tree-gated deep ensemble. In such a case, the weak predictors can each specialize on a certain region of the exogenous representation space, where the intra-class variability is lower. This way, each weak predictor can capture more subtle, cluster-specific, differences between the different classes, resulting in an improved overall accuracy. Note that intuitively this explanation is based on two premises: First, the exogenous variable shall explain a lot of intra-class variability in the data. Second, the exogenous information shall be correctly recognized, \textit{i.e.} a simple predictor placed on top of $h_{exo}$ shall have high accuracy for predicting this exogenous task.

\subsection{Learning disentangled representations with an exogenous dispelling loss}\label{sec:disentangling}

The tree-gated deep ensemble can be trained in an end-to-end fashion by optimizing the following loss over the parameters $\phi$, $\theta$ and $\psi$ the weights of the endogenous representation, classification networks, and tree gate respectively, while $\tau$ the weights of the exogenous network are kept frozen:

\begin{equation}
\mathcal{L}_{sup}(\theta, \phi, \psi, \tau)=CE(y, y^*)
\end{equation}

Where $y=\textbf{Z}(\phi, \theta).\textbf{g}(\tau, \psi)$ and CE denotes the standard cross entropy loss. 

In addition to condition the classification by $\textbf{h}_{exo}$, these features can also be used to disentangle the information related to the exogenous variabilities to the task in the endogenous representation $\textbf{h}_{endo}$. Thus, the main variations not directly related to the task are handled through the tree-gates, so that the weak classifiers decipher the task from their own representation that is invariant to these exogenous variations.
To do this, $\textbf{h}_{exo}$ and $\textbf{h}_{endo}$ should be considered semantically orthogonal. 

Let's assume that the networks $F_{\phi}$ and $\gamma_{\tau}$ (extracting the exogenous and endogenous features respectively) have the same architecture. The extracted features $\textbf{h}_{exo}$ and $\textbf{h}_{endo}$ can then be feed into the exogenous predictor $c_{\sigma}$ to output $\textbf{z}_{exo}^{(exo)}$ and $\textbf{z}_{exo}^{(endo)}$ respectively.
Semantically orthogonalizing the exogenous and endogenous representations therefore means making $\textbf{z}_{exo}^{(endo)}$ as unsimilar as possible to $\textbf{z}_{exo}^{(exo)}$. The similarity $\mathcal{L}_{sim}$ between $\textbf{z}_{exo}^{(exo)}$ and $\textbf{z}_{exo}^{(endo)}$ can be measured by their angular distance:

\begin{equation}
  \mathcal{L}_{sim}(\phi) = \mid cos(\textbf{z}_{exo}^{(exo)}, \textbf{z}_{exo}^{(endo)}) \mid = \frac{\mid \textbf{z}_{exo}^{(exo)}. \textbf{z}_{exo}^{(endo)}\mid } {||\textbf{z}_{exo}^{(exo)}||_2.||\textbf{z}_{exo}^{(endo)}||_2}
\end{equation}

The final loss can then be written as:

\begin{equation}
\mathcal{L}(\theta, \phi, \psi)=\mathcal{L}_{sup}(\theta, \phi, \psi)+\lambda \mathcal{L}_{sim}(\phi)
\end{equation}

With $\lambda \in \mathbb{R}_+$ an hyperparameter whose setting will be discussed in experiments. Having shown how to learn THIN model by applying conditioning w.r.t. an exogenous representation, and using it to remove exogenous information from the endogenous representation, we discuss in the upcoming section the applications of THIN as well as a number of implementation details.

\subsection{Applications and choice of exogenous variables}\label{sec:difer}

\paragraph*{Synthetic datasets} we first apply our THIN model to 2 synthetic datasets where, due to the dataset elaboration process, an exogenous variable can clearly be identified.

\textbf{Digit classification under rotation on the MNIST-Rotated or MNIST-R database}. The database contains the $70k$ images from the well-known handwritten digit MNIST database \cite{lecun1998mnist}, but whose each image is augmented with a random rotation from -90\degree to +90\degree (18 rotation classes, with bins of 10\degree). In this case, we naturally use the rotation as the exogenous variable. All samples are grayscale images that are annotated with 10 different classes, one class per digit, from 0 to 9. Each digit class has approximately the same number of samples. We classically train our models with $60k$ samples and test on $10k$ samples.
\textcolor{black}{We also generate a new version of \textbf{MNIST}, called \textbf{MNIST-RS} augmented both in rotation (with the same setup and classes) and scaling (random scale from 0.5 to 1 divided in 10 classes). }

\textbf{Shape recognition on the dSprites database} \cite{dsprites17} is commonly used for learning disentangled representations. It contains the 2D shapes generated from 6 ground truth independent latent factors. These factors are color, shape, scale, rotation, x and y positions of a sprite. All possible combinations of these factors are present exactly once. We have defined the scale of the shape as the exogenous variable so as to select $60k$ images for training and $10k$ images for testing. Scales are annotated with 10 different classes, by bins of 0.05 from 0.5 to 1.

\paragraph*{Facial expression recognition (FER)} we then apply THIN to real-world FER datasets with identity as the exogenous variable. Facial expression recognition (FER) consists in estimating the perceived emotional state of a person based on facial cues, with applications in human-computer interaction \cite{cowie2001emotion}\cite{bartlett2003real}, virtual/augmented reality \cite{bekele2013understanding}\cite{chen2015augmented}, advanced driver assistance systems \cite{assari2011driver}\cite{jabon2010facial}, education \cite{kapoor2007automatic}, entertainment \cite{lankes2008facial} and healthcare \cite{dapogny2018jemime}.
However, the performance of current FER models is limited in unconstrained conditions containing large variations. We use identity as an exogenous variable because it accounts for a large part of the intra-class variability, as identity encompasses particular morphological traits, gender, age as well as ethnicity.

\begin{table}[]
    \centering
    \caption{Accuracies for the exogenous variables.}
    \begin{tabular}{l|c}
    \hline
    Exogenous variable & Accuracy (\%)\\
    \hline
    Rotation (MNIST-R) & \textcolor{black}{98.32}\\
    Scale (dSprites) & \textcolor{black}{94.18}\\
    Identity (LFW) & 98.95\\
      \hline
    \end{tabular}
    \label{exo_pred}
\end{table}

\paragraph*{Prediction of the exogenous variable} remember that a suitable exogenous variable for our model should explain a lot of intra-class variability (which is clearly the case for our three applications and candidates) and should be predicted with high accuracy. In order to assess this, we report in Table \ref{exo_pred} the accuracies for the exogenous variables that will be used for digit classification, shape recognition as well as for FER, with respect to their respective ranges: \textcolor{black}{for Rotation and Scale, the range is $[-90\degree,90\degree]$ and $[0.5,1]$, respectively. For Identity, we report the pairwise accuracy on LFW database \cite{LFWTech}.} In all cases, these accuracies are very high: Hence, these variables are suitable exogenous variables candidates.

\section{Experiments}\label{sec:experiments_fer}

In this section, we begin (Section \ref{sec:experiments_mnist_dsprites}) by evaluating our model on synthetic datasets, where the exogenous variable can be clearly identified in order to validate the interest of our approach. After this, we validate our model both qualitatively and quantitatively in real-world FER datasets. First, in Section \ref{emo_datasets} we present the datasets that we use to train or test the proposed approach. Then, in Section \ref{implemdetails}, we provide implementation details to ensure reproducibility of the results. In Section \ref{ablation}, we validate the proposed approach by a thorough ablation study of its components. In Section \ref{compsota} we compare THIN with recent FER methods, showing that it significantly outperforms the current state-of-the-art. Last but not least, in Section \ref{modelintro}, we introspect the models to qualitatively assess the interest of our contributions.

\subsection{Evaluation on synthetic datasets}\label{sec:experiments_mnist_dsprites}

\paragraph*{Implementation details} for the benchmarks on MNIST-R, MNIST-RS and dSprites, the baseline model is composed of 2 convolutional layers $ 16 \verb+@+ 3 \times 3$ with max-pooling and ReLU activation. The classification layer consists in 2-FC layers containing a hidden layer of 256 units, also with ReLU activation. Moreover, we apply batch normalization before each activation. For the ensemble models, we use an ensemble of $L=8$ weak predictors. Each of these weak predictors is a $2$-FC layers classifier containing a hidden layer with 32 units each. For the tree-gated deep ensemble models, we use a tree with depth equal to $\log_2(L)=3$. Each model has roughly the same number total number of parameters ($200k$ parameters total). Finally, for the exogenous prediction layer used to compute the dispelling loss $\mathcal{L}_{sim}$, we use a $2$-FC layers classifier containing a hidden layer with 256 units.

Training is done by optimizing the total loss $\mathcal{L}$ over the parameters for endogenous representation $\phi$, classification $\theta$ and gating $\psi$ layers. These weights are optimized jointly in an end-to-end manner by applying ADAM optimizer \cite{kingma2014adam} with a learning rate of $1e$-$3$ and batch size 32.

\paragraph*{Ablation study}\label{toy_ablation}
In this section, we compare the different architectures detailed in sections \ref{sec:deepensemble}. Keep in mind that, to ensure a fair comparison, all models were designed to have roughly the same number of parameters. The ablation study will be conducted in two parts. First, we compare the different architectures. Second, we validate the interest of the proposed dispelling loss.

\paragraph*{Architecture comparison} Table \ref{toy_architectures_comp} showcases a comparison of results obtained with different architectures. First, 
the exogenous conditionned deep ensemble is significantly more robust than the baseline model (+1.28\% on MNIST-R, +1.62\% on dSprites). Conversely, simple and tree-gated deep ensembles do not necessarily perform better than the baseline, which is already efficient in both cases. Likewise, this is due to its single large hidden layer, which alone can encompass all the variability for these relatively simple recognition tasks. 
Nevertheless, conditioning the problem using the exogenous representation, and using a gated ensemble with a set of specialized classifiers on each region of the exogenous representation space increases the overall robustness to the most extreme variations and thus improve the baseline, even with far fewer hidden units by classifier.
Note that the tree-gates allow a better mixture of the classifiers compared to a simple deep ensemble (+3.49\% on dSprites, +0.52\% on MNIST-R), due to the ability to be specialized on a certain region of the gating space. In all cases, the performance of the exogenous tree-gated deep ensemble is very close to the oracle predictor (\textit{i.e.} a deep ensemble tree gated with the ground truth exogenous variable), which may constitute a ceiling in this experiment.

\begin{table}[]
\centering
\caption{Comparison of different architectures in term of average accuracy ($\%$) on MNIST-R and dSprites databases. $^\dagger$: oracle classifier (tree-gated deep ensemble conditionned by the ground truth rotation).}
\resizebox{0.48\textwidth}{!}{
\begin{tabular}{l|c|c}
  \hline
  Method & MNIST-R & dSprites \\
  \hline
  \hline
   Baseline & 96.83 & 96.53 \\
   \hline
    Simple deep ensemble & 96.81 & 92.68 \\
   Tree-gated deep ensemble & 97.31 & 95.91 \\
   Exogenous tree-gated deep ensemble & \underline{98.07} & \underline{98.1} \\
  \hline
   Oracle$^\dagger$ & 98.06 & 98.43 \\
  \hline
  THIN & \textbf{98.26} & \textbf{98.5} \\
  \hline
\end{tabular}}
\label{toy_architectures_comp}
\end{table}

\begin{figure}
    \centering
    \includegraphics[width=\linewidth]{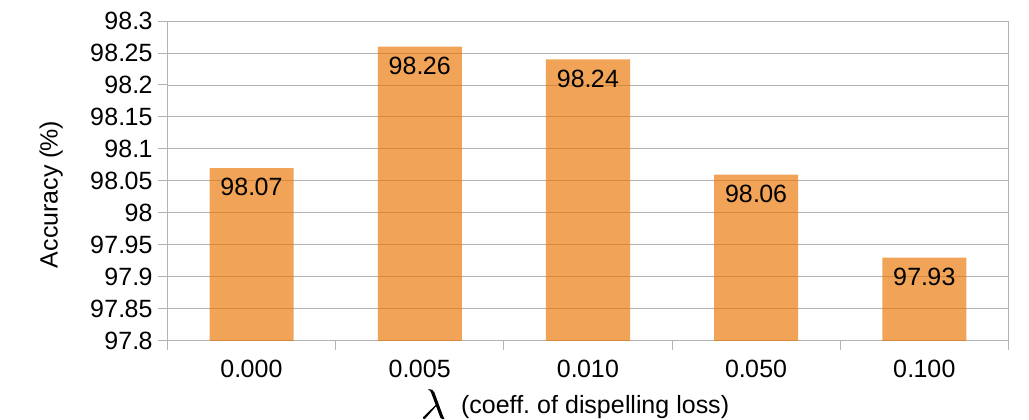}
    \caption{Ablation study for the exogenous representation-dispelling loss in term of average accuracy ($\%$) on MNIST-R.}
    \label{toy_ablation_study}
\end{figure}

\textcolor{black}{\paragraph*{Handling several factors of variation} 
Table \ref{several_exo_var} showcases a comparison of tree-gated deep ensemble methods in term of average accuracy on MNIST-RS, depending on the different gating variable chosen: the digit endogenous representation, the rotation representation, the scale representation, or their concatenation. First, we observe that either rotation and scale allow to better condition digit recognition compared to the digit endogenous representation (+0.56\% for the scale, +0.74\% for the rotation), leading to increase the robustness to their respective variations, thus improving the overall accuracy. Second, rotation acts as a better exogenous variable than scale due to its dramatic influence on digit appearance, leading to better condition digit recognition and increase the overall accuracy. Finally, using both rotation and scale allows to encompass more exogenous variations, leading to further improve the digit recognition and the overall accuracy.}

\begin{table}[]
\centering
\caption{\textcolor{black}{Accuracies for the tree-gated deep ensemble methods in term of average accuracy (\%) on MNIST-RS, depending on the gating variable chosen.}}
\begin{tabular}{l|c}
  \hline
  Gating variable & MNIST-RS \\
  \hline
  \hline
   Digit & 96.68 \\
   \hline
   Scale & 97.22 \\
   Rotation & \underline{97.4} \\
   Rotation + Scale & \textbf{97.63} \\
  \hline
\end{tabular}
\label{several_exo_var}
\end{table}

\paragraph*{Dispelling loss} Figure \ref{toy_ablation_study} show the accuracies on MNIST-R with different values of the coefficient $\lambda$ in the dispelling loss $\mathcal{L}_{sim}$ at train time. 
A particular care must be taken for tuning $\lambda$ whose optimal value depends on the task carried out. Indeed, if $\lambda$ is too strong, the training favors the dispelling of the exogenous information from the endogenous representation rather than maximizing its discriminatory power, hence lowering its accuracy.
Thus, on MNIST-R, if we set $\lambda=0.005$, the digit classification and rotation-dispelling from digit features are more balanced. By removing undesirable rotation-related variations in the digit representation, accuracy is improved by 0.2\%.
This is further confirmed on dSprites where an improvement of 0.41\% can be observed by removing scale-related variations in the shape representation.

\subsection{FER Datasets}\label{emo_datasets}

We thoroughly validate our approach on the most recent and challenging FER databases: RAF-DB, AffectNet and ExpW.

The \textbf{Real-world Affect database or RAF-DB} \cite{li2017reliable} contains $30k$ facial images annotated with basic or compound expressions. RAF-DB contains a great diversity in identity: 52\% female, 43\% male, 5\% unsure, from 0 to 70 years old, and the ethnicity distribution is 77\% Caucasian, 8 \% African-American, and 15\% Asian. These images have been annotated by 315 human coders. Each image has been annotated 40 times, and the final annotation was obtained \textit{via} crowdsourcing methods. As it is traditionally done in the literature, we only used the aligned images labelled with basic facial expressions, which makes a total of $12271$ examples for train/validation and $3068$ examples for test.

The \textbf{AffectNet} database \cite{mollahosseini2017affectnet} is the largest labelled FER database, and contains $400k$ images manually annotated with basic facial expressions as well as valence/arousal intensities. These images have been collected by querying three major search engines using 1250 expression related keywords in six different languages. Similarly to what is done in the literature, we only used the images with basic FEs, allowing us to train the models with $280k$ samples and test on $3500$ samples. We used the face alignment model \cite{ArnaudFG2019} to localize facial landmarks for automatically cropping each image. The models are then trained with alternated sampling between each FE class for each mini-batch.

The \textbf{Expression in-the-Wild or ExpW} \cite{zhang2018facial} is the most recent FER database, containing 91793 faces manually annotated with basic facial expressions. These images were collected using the Google search image API, and a confidence face level is available for each image so that non-image faces can be removed. As is done in the literature, we select images with the confidence face greater than 60, allowing to use 26701 images for training and 6673 images for testing.

\subsection{Implementation details}\label{implemdetails}

\paragraph*{Architectural choices} for the baseline architecture, we employ a VGG-16 architecture: thus, the representation layer is composed of 13 convolutional layers with ReLU activation. For the representation layer we use weights pre-trained either on ImageNet (denoted as VGG16) or on VGG-Face database \textcolor{black}{\cite{Parkhi15}} (denoted as VGGFace). The classification layer is trained from scratch and consists of 3 fully-connected layers, also with ReLU activation. Contrary to the original version, we apply batch normalization before each activation.

For the ensemble models, \textcolor{black}{we set $L$ the number of weak predictors to 32, which performed the best in our experiments}. Each of these weak predictors contains 2 hidden layers with 512 units each. For the tree-gated, exogenous tree-gated as well as THIN models, we use a tree with depth equal to $\log_2(L)=5$. Each model has roughly the same number total number of parameters (100M parameters total).

To model identity, we use a VGG16 network pretrained for face recognition \cite{Parkhi15}. It consists in 13 convolution layers to extract $512$ feature maps $7 \times 7$ for the representation layer $\gamma_\tau$, then 3 fully-connected (FC) layers containing two hidden layers with 4096 units each for the identity-classifier $c_\sigma$. The parameters $\tau$ and $\sigma$ are kept frozen during the training of our FER model.

\paragraph*{Learning THIN}
Training is done by optimizing the total loss $\mathcal{L}$ over the parameters for facial expression representation $\phi$, classification $\theta$ and gating $\psi$ layers. These weights are optimized jointly in an end-to-end manner by applying ADAM optimizer \cite{kingma2014adam} with a learning rate of $1e$-$5$ and batches size 16. For each experiment, we construct a validation set using 256 examples from the training set and select the best model in terms of overall accuracy on this validation set after 100k iterations, and report the accuracy of this model on the test set. Preprocessing of the images include resizing to $224 \times 224$ and applying data augmentation as it is traditionally done in the literature: rotation $\theta \sim Uniform([-10^{\circ}, +10^{\circ}])$, random horizontal flip as well as random brightness, saturation, hue and contrast variations.

\subsection{Ablation study}\label{ablation}

\begin{table}[!t]
\centering
\caption{Comparison of different architectures in term of average accuracy ($\%$) on RAF-DB, AffectNet and  ExpW databases.}
\resizebox{0.48\textwidth}{!}{
\begin{tabular}{l|c|c|c}
  \hline
  Method & RAF-DB & AffectNet & ExpW \\
  \hline
  \hline
   Baseline (VGG16) & 82.99 & 61.31 & 70.96 \\
   Baseline (VGGFace) & 84.06 & 61.66 & 71.57 \\
   \hline
    Simple deep ensemble & 85.59 & 63.00 & 75.05 \\
   Tree-gated deep ensemble & 86.38 & 63.34 & 75.17 \\
   Exogenous tree-gated deep ensemble & \underline{87.29} & \underline{63.71} & \underline{75.74} \\
  \hline
  THIN & \textbf{87.81} & \textbf{63.97} & \textbf{76.08} \\
  \hline
\end{tabular}}
\label{architectures_comparison}
\end{table}

\begin{figure}[!t]
    \centering
    \includegraphics[width=1.\linewidth]{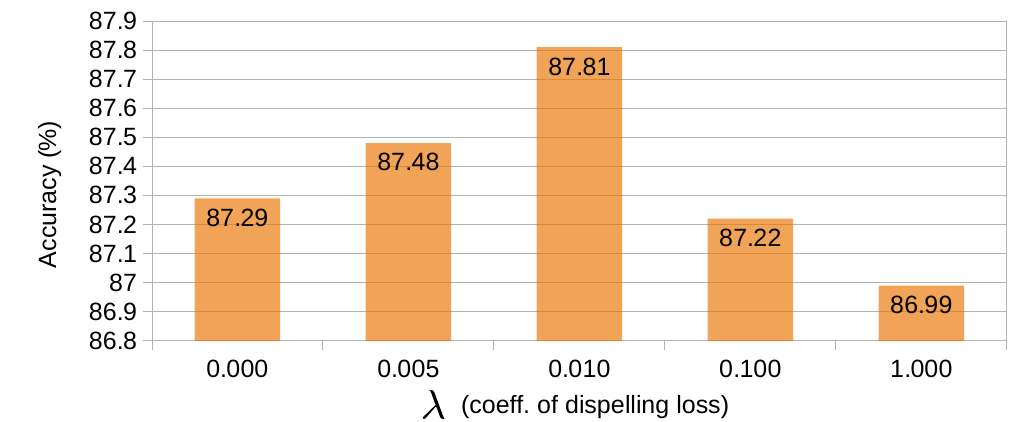}
    \caption{Ablation study for identity-expression disentangling in term of average accuracy ($\%$) on RAF-DB.}
    \label{ablation_study}
\end{figure}

In this section, we compare the different architectures detailed in sections \ref{sec:deepensemble}. Here again, all models were designed to have roughly the same number of parameters.

\paragraph*{Architecture comparison} Table \ref{architectures_comparison} showcases a comparison of results obtain with different architectures. Firstly, representations pretrained for face recognition (VGGFace) rather than on ImageNet (VGG16) are better suited for deciphering FEs ($+1.29\%$ on RAF-DB, $+0.57\%$ on AffectNet,$+0.86\%$ on ExpW). This is natural since, even though FER and identity classification are semantically orthogonal tasks, the domain gap is less important as compared to object classification on datasets such as ImageNet.

Secondly, the simple deep ensemble is significantly more robust than the baseline model, on all databases ($+1.82\%$ on RAF-DB, $+2.17\%$ on AffectNet, $+4.86\%$ on ExpW). This can be explained by the fact that, for tasks such as FER where there is a large intra-class variation \cite{cai2018island}, using ensemble learning allows to learn a set of more specialized classifiers, which, in turn, enhances the inter-class/intra-class ratio. Indeed, this ratio is equal to $0.27$/$0.24$/$0.20$ on RAF-DB/AffectNet/ExpW respectively for the baseline network, and $0.44$/$0.34$/$0.40$ for the single deep ensemble.

Furthermore, adding the tree gate to design better mixtures of weak predictors allows each of these weak predictors to be more specialized towards a certain region of the input gating embedding space. Because of this, the tree-gated deep ensemble is slightly more robust on ExpW ($+0.16\%$), on AffectNet ($+0.54\%$), and more so on RAF-DB ($+0.92\%$).

Lastly, using exogenous (identity) representations as the tree gate input rather than expression representation leads to enhanced performance. As such, using identity as the input to the tree gates, the accuracy again increases by $+1.05\%$ on RAF-DB, $+0.76\%$ on ExpW and $+0.58\%$ on AffectNet. This validates our claim that, as a variable explaining a lot of intra-class variability, identity appears as an overall better suited conditioning variable.

\paragraph*{Identity-dispelling loss} next, in Figure \ref{ablation_study} we show the variation of the FER accuracy on RAF-DB with different values of the coefficient $\lambda$ of the identity-dispelling loss $\mathcal{L}_{sim}$ at train time. First, $\lambda$ should be set with care in order to avoid that $\mathcal{L}_{sim}$ overrides the FER loss $\mathcal{L}_{sup}$ ($\lambda>0.1$). With this in mind, e.g. if we set $\lambda=0.005$ and, \textit{a fortiori} $\lambda=0.01$ the FE classification and identity-expression orthogonalization terms are more balanced. In such a case, the proposed identity-dispelling loss allows to significantly enhance the FER accuracy ($+0.60\%$ on RAF-DB, $+0.41\%$ on AffectNet, $+0.45\%$ on ExpW), by removing undesirable identity-related variations in the expression representations. This also echoes the qualitative assessments that will be introduced in Section \ref{modelintro}.

\subsection{Comparison with state-of-the-art approaches}\label{compsota}

\begin{table}[]
\centering
\caption{Comparison with state-of-the-art approaches in term of accuracy ($\%$).}
\begin{tabular}{l|c|c|c}
  \hline
  Method & RAF-DB & AffectNet & ExpW\\
  \hline
  \hline
   PG-CNN \cite{Li2018PatchGatedCF} & 83.27 & 55.33 & -\\
   Separate loss \cite{li2019separate} & 86.38 & 58.89 & -\\
   IPA2LT \cite{Zeng2018FacialER} & 86.77 & 57.31 & -\\
   RAN \cite{wang2019region} & 86.9 & 59.5 & -\\
   Covariance pooling \cite{Acharya_2018_CVPR_Workshops} & \underline{87.00} & - & -\\
   SNA \cite{fu2020semantic} & - & 62.7 & - \\
   BReG-Net \cite{hasani2019bounded} & - & \underline{63.54}&  -\\
   PAT-VGG \cite{cai2019improving} & 86.28 & - & 71.5 \\
   EAFR \cite{lian2020expression} & 82.69 & - & \underline{71.90} \\
  \hline
   THIN & \textbf{87.81} & \textbf{63.97} & \textbf{76.08} \\
  \hline
\end{tabular}
\label{sota}
\end{table}

\begin{figure*}
    \centering
    \includegraphics[width=\linewidth]{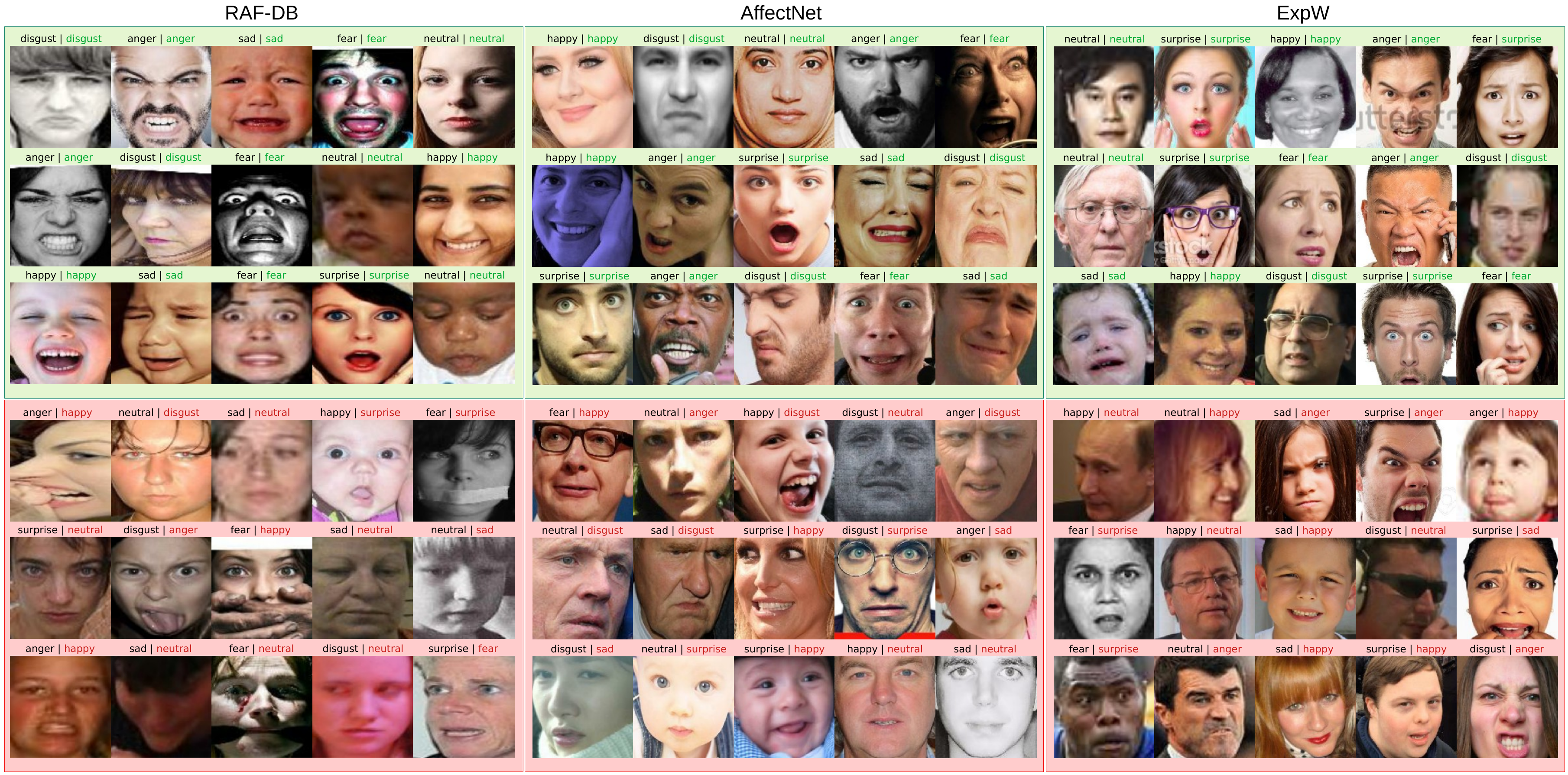}
    \caption{Examples of good (green) and bad (red) predictions on all three FER datasets. For each example, the ground truth FE class is overlaid on the image (in black) with the corresponding prediction (in green or red). Note that misclassifications can only be due to wrong annotations, especially on AffectNet and ExpW whose some (webscrapped) images have not been or badly reannotated by humans. While the main characteristics are present (e.g. presence of a smile, wrinkled eyes for the happy expression), the expression can be badly annotated but well classified by our model.}
    \label{fig:goodbad}
\end{figure*}

\begin{figure*}
    \centering
    \includegraphics[width=\linewidth]{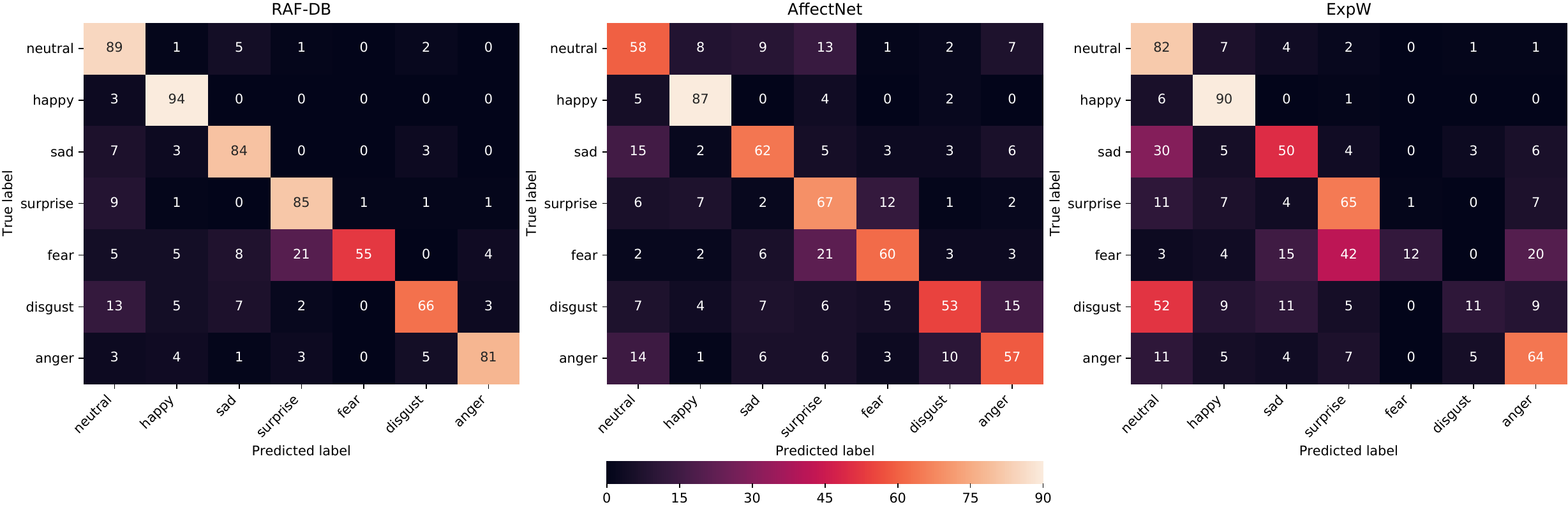}
    \caption{Confusion matrices (\%). On RAF-DB, AffectNet and ExpW testsets, THIN accurately predicts most expression classes.}
    \label{fig:confusion_matrix}
\end{figure*}

Table \ref{sota} displays a comparison between THIN and other recent state-of-the-art methods on RAF-DB, AffectNet and ExpW databases. THIN significantly extends the state-of-the-art over these approaches, as top accuracy increases from 87.0\% to 87.81\% on RAF-DB, and from 63.54\% to 63.97\% on AffectNet. This is likely due to the fact that that, contrary to its closest contenders (RAN \cite{wang2019region} and Covariance pooling \cite{Acharya_2018_CVPR_Workshops} on RAF-DB, and BReG-Net \cite{hasani2019bounded} on AffectNet), THIN explicitly leverages identity information to enhance FE classification and representation in two different ways: on the one hand, by using identity as a gating variable to adaptively weight the deep ensemble mixture and, on the other hand, by explicitly removing identity information from the expression representation \textit{via} the identity-dispelling loss. Therefore, THIN effectively reduce the variations related to morphological traits, which is crucial to designing efficient FER systems that generalize well on large numbers of identities. On AffectNet, it should be noted that according to \cite{mollahosseini2017affectnet}, the annotator's agreement is $65.3\%$, thus this might constitute a ceiling in performance, from which our method allows to substantially narrow the gap.

Figure \ref{fig:confusion_matrix} shows the confusion matrices for THIN on RAF-DB, AffectNet and ExpW. Generally speaking, happiness is the most accurately predicted expression class, then neutral, sadness, anger and surprise, while fear and disgust are the least accurate classes. On RAF-DB and more importantly on ExpW, the prediction appears to be biased towards the neutral class. This may indicate that some sort of batch balancing could be implemented to output more balanced predictions.

However, it should be noted that the quality of annotations on the FER datasets, especially on AffectNet and ExpW, is not optimal because some webscrapped images have not been or badly reannotated by humans. Figure \ref{fig:goodbad} shows examples of good and bad classifications on all three FER datasets: it can then be seen that some obvious expressions, whose main characteristics are well present (\textit{e.g.} presence of a smile and squinted eyes for the happy expression), have been badly annotated but well classified by our model.

\subsection{Model introspection}\label{modelintro}
In this section, we provide insight on the underlying mechanisms behind THIN behavior. Namely, we show that, to a certain extent, both the proposed identity tree-gating scheme and identity-dispelling loss acts by effectively pushing away identity information from the expression-relation representations. \textcolor{black}{Second, we show that this allows a better clustering of the distinct FEs in this representation space. Finally, we visualize the regions of the identity space upon which each FE classifier is specialized.}

\begin{figure}
  \centering
  \includegraphics[width=1.\linewidth]{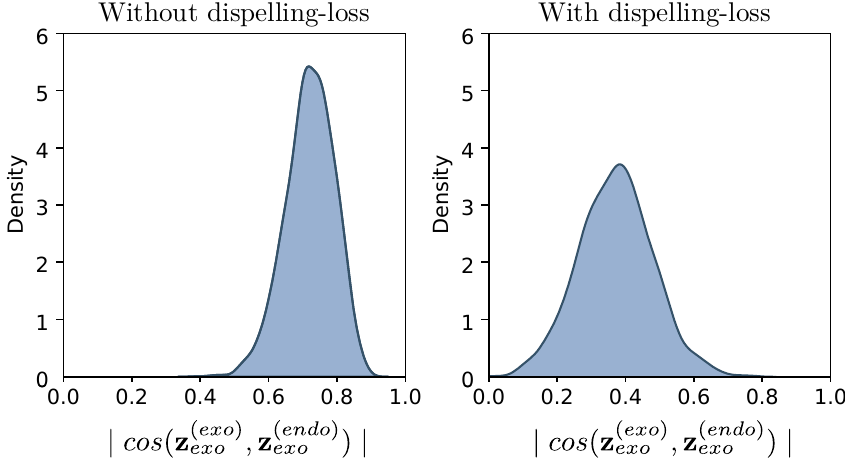}
  \caption{Distributions of absolute cosine distances $\mathcal{L}_{sim}(\phi)$ on RAF-DB. Left: $\lambda=0$ (no dispelling). Right: $\lambda=0.01$. The identity-dispelling loss effectively orthogonalizes identity and expression representations.}
  \label{cos}
\end{figure}

\paragraph*{Dispelling identity in expression representation} figure \ref{cos} shows the distribution of absolute cosine distances with and without the identity-dispelling loss $\mathcal{L}_{sim}$ during training. We observe that the angles with $\lambda>0$ are significantly closer to 0, which indicates more orthogonality between the identity and expression representations.

\begin{figure}
  \centering
  \includegraphics[width=\linewidth]{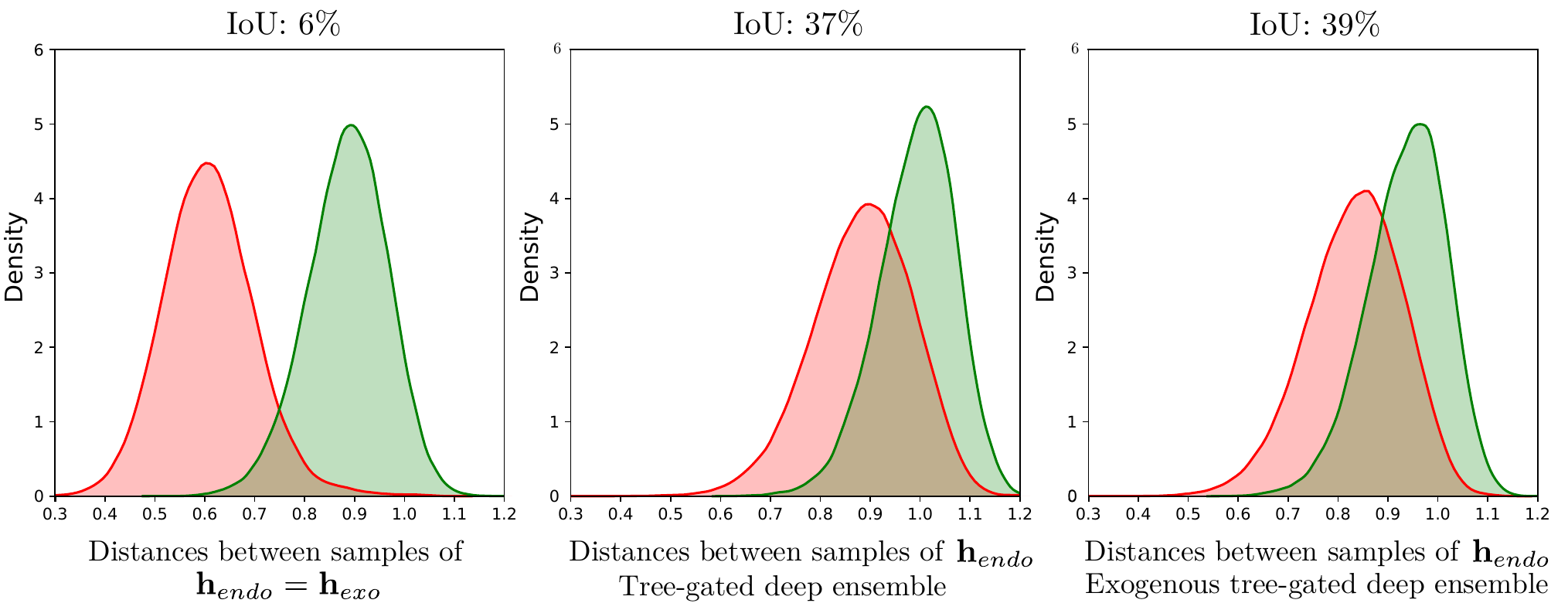}
  \caption{Distance distributions of the endogenous representations between according to positive/negative sample pairs from LFW \cite{LFWTech}. In red: same identity between samples. In green: different identities between samples. The left figure corresponds to the training initialization of the endogenous representation with identity features: it indicates the case where the identities are separated as much as possible.}
  \label{lfw_1}
\end{figure}

In the next experiment, we use the test partition of the Labeled Faces in the Wild (LFW) \cite{LFWTech} database, which is frequently used for benchmarking face recognition systems. Specifically, we generate a balanced mix of positive (\textit{i.e.} which belongs to the same identity) and negative pairs (\textit{i.e.} which belongs to different identities) and plot the distribution of the distances within the ($\mathcal{L}_2$-normalized) representation spaces. The results are shown on Figure \ref{lfw_1}. The red curve is associated with the positive pairs and the green curve is associated with the negative ones.

First, as can be seen on the left graph, in the identity representation space, the positive and negative pairs can be easily distinguished, e.g. by applying a threshold at $0.75$ to the normalized representation distance. In such a case, the two distributions do not overlap much, as also indicated by the low Intersection-over-Union (IoU) score (on top of the graphs on Figure \ref{lfw_1}): the identity representation naturally appears as a suitable space for discriminating different identities. Conversely, for the expression representation space (e.g. for the tree-gated deep ensemble-central graph) the two distributions overlap to a significant extent: as identity and expression recognition are semantically orthogonal tasks, a representation has to lose its identity-discriminative capacity in order to become a suitable expression representation. The plot for the Exogenous tree-gated deep ensemble (right graph) is very similar but with a slightly higher IoU between the two distributions: in the tree-gated deep ensemble, the expression representation has to store some information related to faces morphological traits to suitably condition the weak predictor choice. By contrast, with the Exogenous tree-gated deep ensemble, this information can be stored in the identity representation, leaving more room for the expression-related information.

\begin{figure*}
  \centering
  \includegraphics[width=\linewidth]{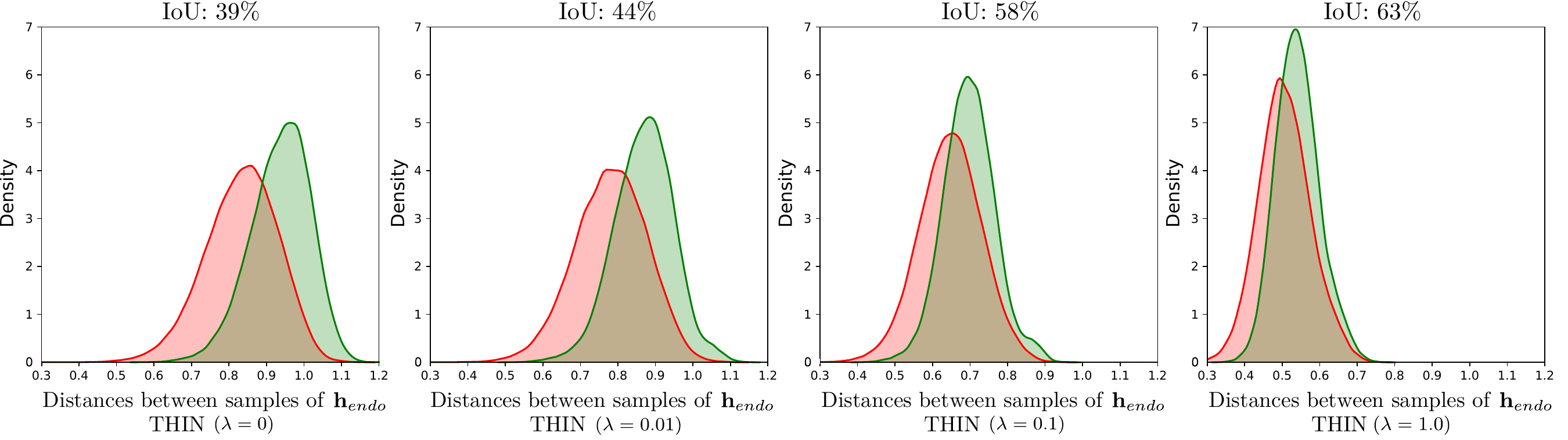}
  \caption{Distance distributions between according to positive/negative sample pairs from LFW \cite{LFWTech} for THIN models. In red: same identity between samples. In green: different identities between samples.}
  \label{lfw_2}
\end{figure*}

Figure \ref{lfw_2} shows the same positive/negative distance distribution for THIN models. The identity-dispelling loss increases the IoU between the two distributions, thus helping the expression representation to forget all the information related with morphological traits. However, with $\lambda>0.1$ the main focus of the expression representation becomes to be bad at identity recognition, not to be good at expression prediction. Hence, in such a case, not enough emphasis can be put on the FER task, which can lead to a decreased performance.

\begin{figure}
    \centering
    \includegraphics[width=\linewidth]{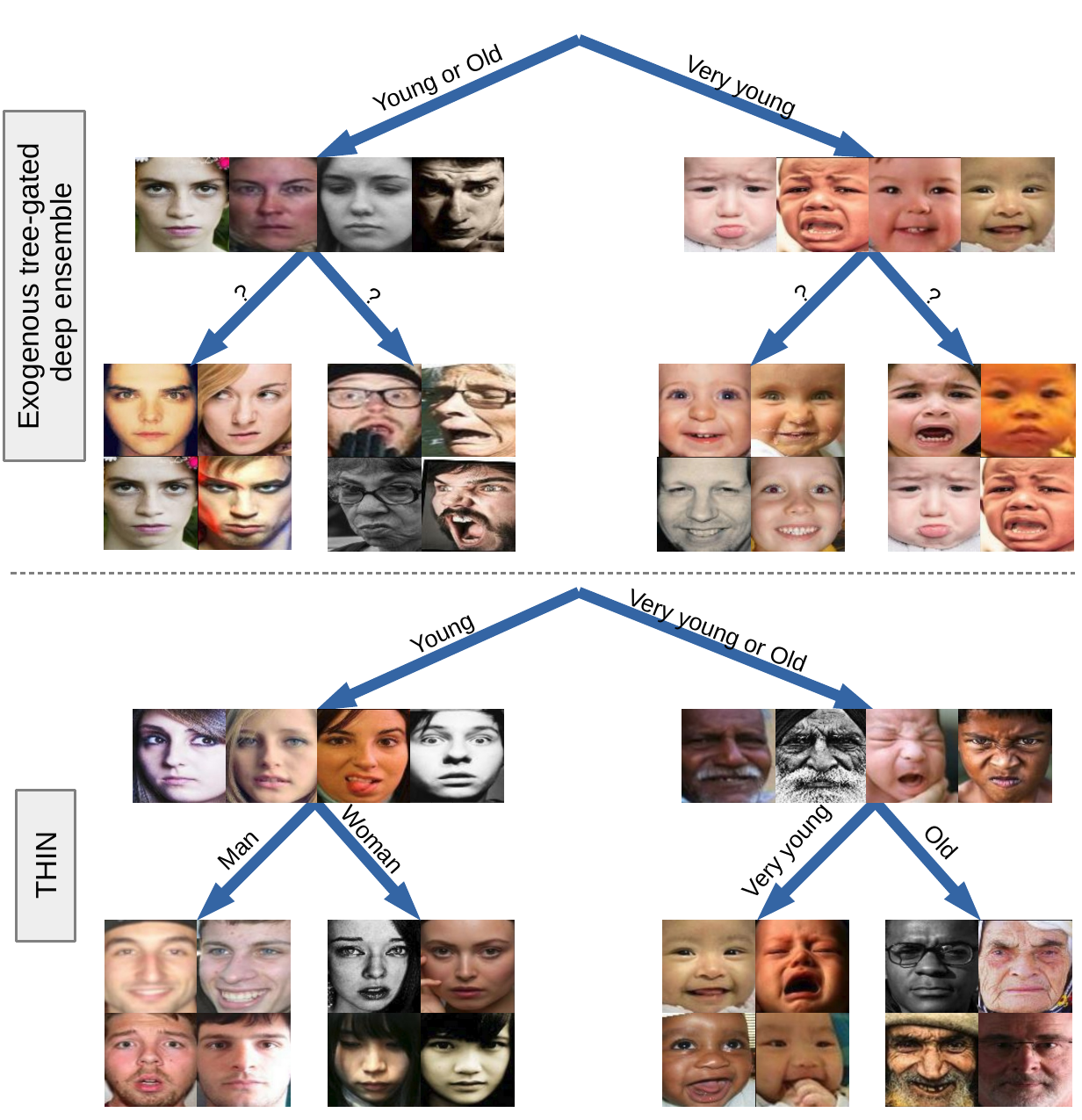}
    \caption{\textcolor{black}{Comparison of the hierarchical partitions of the identity (exogenous) representation space learned by using the dispelling loss or not. Images from the RAF-DB testset.}}
    \label{tree_vis}
\end{figure}

\begin{figure*}
  \centering
  \includegraphics[width=\linewidth]{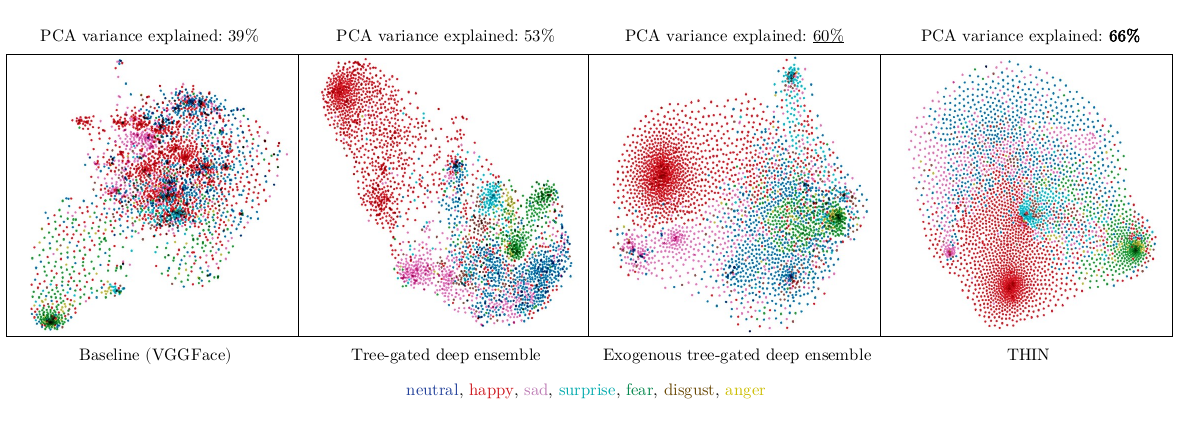}
  \caption{Visualization of expression features on RAF-DB (dimension reduction with t-SNE algorithm). At the top of each figure, the percentage of variance explained on the first three axes of the PCA.}
  \label{tsne}
\end{figure*}

\paragraph*{Expression features distribution} we use t-SNE \cite{maaten2008visualizing} to visualize the clustering of images with different FE classes on RAF-DB in a low-dimensional projection of the expression space. The results are illustrated on Figure \ref{tsne}. First, we observe that the tree-gated deep ensemble enables a clearer repartition of the different FE classes in the expression space, Moreover, using identity representation as the gate input also allows a better repartition of the various FE classes, e.g. for expressions sadness, and surprise. Last but not least, THIN (with the addition of the identity-dispelling loss) provide a better separation of expressions anger, fear and surprise, leading to an increased overall performance.

\textcolor{black}{\paragraph*{Partition of the identity space}
Regarding the exogenous representation space, we visualize the regions upon which each weak classifier is specialized. Figure \ref{tree_vis} shows the most representative face images on the RAF-DB testset (\textit{i.e.} the images with the highest probability to reach each split node) for each node of the tree-gates defining a partition in identity space, depending on whether the dispelling loss is used during training (bottom) or not (top). First, we observe that, for the exogenous tree-gated deep ensemble, we observe that the tree-gate learns to partition the identity space between \textit{young/old} and \textit{very young} persons. However, the learned partition is not clear-cut beyond the first split node. Second, for THIN, the regions seem then to be more visually interpretable thanks to the proposed dispelling loss: as we encourage the removal of exogenous information (\textit{i.e.} the identity representation) from endogenous representation (\textit{i.e.} the FE representation), this information can then be better leveraged by gates, which learn more efficient regions, in turn leading to better specialized weak classifiers and enhanced overall accuracy. In such a case, we observe that the identity space is clearly divided in terms of age, then in terms of gender.}

\section{Conclusion}\label{sec:conclusion}
Throughout this work, we studied a special case of computer vision task where an exogenous variable can be identified, such that (a) this variable shall explain a lot of intra-class variability, and (b) the task prediction shall be invariant to this variable. To address this kind of task, we model a dual representation: an exogenous representation and an endogenous representation. We propose an exogenous tree-gated deep ensemble that employs a differential tree gate that learns to adaptively weight the weak predictors, depending on the exogenous representation, explicitly modeling the dependency between the exogenous variable and the predicted task (a), and ultimately resulting in better performance.
Furthermore, we propose an exogenous dispelling loss that removes the exogenous information from the endogenous representation, enforcing (b). We call this method THIN, standing for THrowable Information Networks. We experimentally demonstrate that, provided an exogenous information can be identified, THIN substantially improves the prediction accuracy on a variety on benchmarks, including synthetic data (rotated digit classification, with rotation as the exogenous variable or shape recognition with shape scale as the exogenous variable). We also apply it to FER in-the-wild with identity as the exogenous variable, and perform extensive experiments on several challenging datasets. Specifically, we show that THIN significantly outperforms existing state-of-the-art FER approaches.

The proposed approach opens up a lot of interesting directions in a variety of computer vision and deep learning problems. First, THIN could in theory be applied quite straightforwardly to other problems, for instance body pose estimation, with the orientation or scale as the exogenous variable, or semantic segmentation with domain information (e.g. the nature of the landscape-indoor/outdoor environment, urban scenes, and so on) as the exogenous variable. Second, in this paper, we only used one exogenous variable to train THIN. However, we could easily imagine using multiple such variables and representation networks with some kind of fusion scheme to apply conditioning w.r.t. these multiple exogenous representations. For instance in the case of FER, we could use identity, as well as environmental lighting, and head pose as exogenous variables. Furthermore, if we consider identity as exogenous for FER, the reciprocal is also true, meaning that FEs are also exogenous to identity recognition. Thus, we could use THIN with identity as the exogenous variable to predict FEs, then use another THIN to predict identity using FE as the exogenous variable, and so on, to iteratively refine both FER and identity prediction.

\section*{Acknowledgment}
This work has been supported by the French National Agency (ANR) in the frame of its Technological Research JCJC program (FacIL, project ANR-17-CE33-0002).

\bibliographystyle{IEEEtran}
\bibliography{references}

\end{document}